# Deep Learning for Unsupervised Anomaly Localization in Industrial Images: A Survey

Xian Tao, *Member, IEEE,* Xinyi Gong, Xin Zhang, Shaohua Yan, and Chandranath Adak, *Senior Member, IEEE*

*Abstract*—**Currently, deep learning-based visual inspection has been highly successful with the help of supervised learning methods. However, in real industrial scenarios, the scarcity of defect samples, the cost of annotation, and the lack of a priori knowledge of defects may render supervised-based methods ineffective. In recent years, unsupervised anomaly localization algorithms have become more widely used in industrial inspection tasks. This paper aims to help researchers in this field by comprehensively surveying recent achievements in unsupervised anomaly localization in industrial images using deep learning. The survey reviews more than 120 significant publications covering different aspects of anomaly localization, mainly covering various concepts, challenges, taxonomies, benchmark datasets, and quantitative performance comparisons of the methods reviewed. In reviewing the achievements to date, this paper provides detailed predictions and analysis of several future research directions. This review provides detailed technical information for researchers interested in industrial anomaly localization and who wish to apply it to the localization of anomalies in other fields.**

## I. INTRODUCTION

AUTOMATED visual inspection based on deep learning technology is being widely applied in industrial defect detection applications due to its efficiency and remarkable accuracy, including unmanned aerial vehicle (UAV) patrol inspection of power equipment [1], weak scratches detection on industrial surfaces [2], identification of copper wire defect in deep hole parts [3], conductive particle detection for chip on glass [4] and so on. Existing inspection systems are primarily based on the supervised learning method, which significantly relies on labeled data. Image category labels, bounding box labels, and fine-grain pixel-wise labels are the three classical types of labels available. Unfortunately, the above-mentioned fully-supervised approaches suffer from several inevitable limitations: (i) The ample annotations are labor-intensive and high cost. (ii) As the process on several precision generation lines improves, defective samples are becoming scarce, posing labeling challenges. (iii) All possible defective types need to be known in advance under fully-supervised learning. (iv) Annotation noise may be inadvertently introduced when labeling the data. As a result, both academia and industry have paid extensive attention to developing unsupervised technology for vision inspection systems.

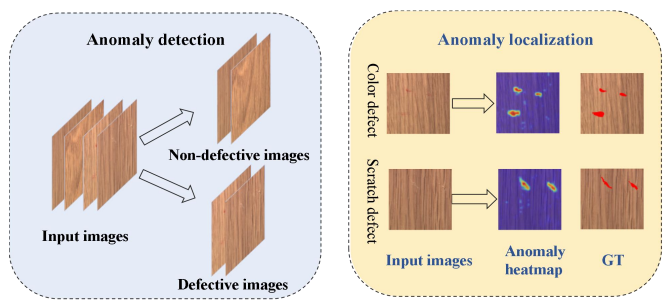

Fig. 1. Anomaly localization vs. Anomaly detection (the samples are from the wood dataset of MVTec AD [115]).

### A. Anomaly Detection versus Anomaly Localization

The human visual system has the inherent ability to perceive anomalies – not only can humans distinguish between defective and non-defective images, even if they have never seen any defective samples before, but they can also point out the location of anomalies. Anomaly localization (AL) was introduced to academia for the very same purpose, *i.e.*, to teach the machine to 'find' the anomaly region in an unsupervised manner. In the context of deep learning methods, 'unsupervised' means that the training stage contains only normal images without any defective samples. AL method under the unsupervised paradigm firstly avoids the hardship of collecting anomalous or defective samples, which cannot be avoided in the supervised method; since the normal images without defects are far more than the abnormal samples in the industrial

This work was supported by the Beijing Municipal Natural Science Foundation (China) 4212044 and the National Natural Science Foundation of China under Grant 62066004. (Corresponding authors: Xian Tao; Chandranath Adak)

Xian Tao, Xinyi Gong, and Shaohua Yan are with the Research Center of Precision Sensing and Control, Institute of Automation, Chinese Academy of Sciences, Beijing 100190, China, and also with the School of Artificial Intelligence, University of Chinese Academy of Sciences, Beijing 100049, China (e-mail: taoxian2013@ia.ac.cn)

Xin Zhang is with Key Laboratory of Industrial Internet and Big Data, China National Light Industry, Beijing Technology and Business University, Beijing 100048, China

Chandranath Adak is with the Dept. of CSE, Indian Institute of Technology Patna, Bihar - 801106, India. (e-mail: chandranath@iitp.ac.in)

TABLE 1 Summary of previous reviews

| Title | Year | Venue | Description |
|---|---|---|---|
| Image Anomalies: A Review and Synthesis of Detection Methods [6] | 2019 | JMIV | This paper reviews the classical image AD models before 2018 and compares 6 representative algorithms on a synthetic database. |
| Deep learning for anomaly detection: A survey [7] | 2019 | Arxiv | This paper reviews deep learning-based AD methods along with their application across various domains. |
| A Unifying Review of Deep and Shallow Anomaly Detection [5] | 2020 | Proc. IEEE | This paper established a systematic unifying view of deep and shallow AD models and discussed many practical aspects. |
| Image/Video Deep Anomaly Detection: A Survey [8] | 2021 | Arxiv | This paper conducts an in-depth investigation into the images/videos of deep learning-based AD methods and discusses challenges and future research directions. |
| Deep learning for anomaly detection: A review [9] | 2021 | ACMCS | This paper surveys deep AD with a comprehensive taxonomy, covering advancements in 3 categories and 11 fine-grained categories of the methods. |
| Visual Anomaly Detection for Images: A Systematic Survey [10] | 2022 | Procedia CS | This paper provides a short survey of the classical and deep learning-based approaches for visual AD and AL. |
| GAN-based Anomaly Detection: A Review [11] | 2022 | Neurocomputing | This paper focus on GAN-based AD and discusses its theoretical basis and applications. |

scenario. Secondly, the labeling cost of the training sample in the supervised method can be eliminated in the unsupervised method. Last but not least, the unsupervised method also avoids the influence of labeling deviation, which is commonly seen in the supervised method. Since the training data only have the normal class, it may be called 'semi-supervised'. However, to unify with most of the current methods, we remove the term 'unsupervised' or 'semi-supervised' in the following content and only call it AL. The distinction between AD and AL is depicted in Fig. 1. Outlier detection or one-class classification are other terms for AD. It refers to the task of distinguishing defective images at the image level from the majority of non-defective images. AL, on the other hand, is also known as anomaly segmentation, which is used to produce pixel-level anomaly location results. The darker the color in the anomaly heat-map, as shown in Fig. 1, the more likely the location is to be anomalous.

The AD task is insufficient to ensure that the method can identify the actual defect locations in real-world industrial scenarios. As AD only performs a binary classification of the image, the results are uninterpretable. Although the image is classified as a defect catalog, the focus areas of the network may not be abnormal. As illustrated in Fig. 2, the anomaly detection method tends to place a high value on wood strains rather than on drill holes that are the real anomalies. Finding anomalies on industrial scene images is the starting point for this survey.

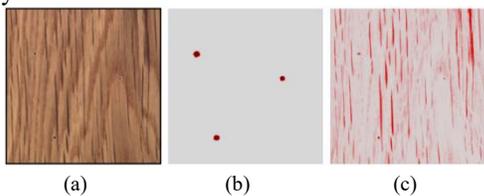

Fig. 2. Apparent limitations of anomaly detection: (a) input image, (b) ground-truth of anomaly (three drill holes), and (c) visual explanation of the anomaly prediction. The model assigns high relevance to the normal texture of wood instead of the drill holes.

### B.  Differences from Previous Surveys

Table 1 enlists existing surveys that are similar to ours. Ehret et al. [6] reviewed classical approaches up to 2018 and thus did not include recent deep learning-based solutions. Wei et al. [7] investigated deep AD in supervised, semi-supervised, and unsupervised domains. Ruff et al. [5] provide a recent comprehensive review of the connections between traditional "shallow" and new "deep" approaches to AD. Mohammadi et al. [8] provided an overview and classification of image/video deep learning-based AD methods, dividing them into three categories: self-supervised learning, generative networks, and anomaly generation. Xie et al. [9] provided a comprehensive taxonomy for deep AD, covering advances in methods for three categories and 11 refinement categories. Yang et al. [10] presented a concise overview of traditional and deep learning-based visual AD and AL techniques. Xia et al. [11] conducted comprehensive survey research on generative adversarial networks (GAN)-based AD.

Multiple surveys related to AD/AL are presented in Table 1, involving studies in areas of early non-deep learning AD methods [6], deep crude AD methods [5, 7-9], limited AL models [10], or focusing only on GAN [11]. However, few surveys are dedicated to comprehensive AL methods. On the other hand, most of the existing surveys cast existing approaches into AD methods for image-level classification. As shown in Fig. 2, major AD methods easily ignore abnormal regions in industrial scenarios. Moreover, in recent five years, AL methods have developed from the image-level comparison (reconstruction or generation) to feature-level comparison, and also from the simple proxy task of defect synthesis to the self-supervised method based on contrast learning. While AL on images or videos has been widely concerned, to the best of our knowledge; still no published paper has summarized detailed improvement and trends, e.g., network structure, loss function, feature comparison method, sample synthesis approach, etc. Our work systematically and comprehensively reviews recent advances in unsupervised AL. It includes an in-depth analysis and discussion of numerous aspects that have never been explored in this area before, to the best of our knowledge. In particular, we summarize and discuss existing methods for deep AL tackling various problems and challenges, provide a roadmap and taxonomy, review the existing datasets and evaluation metrics, present a comprehensive performance comparison of state-of-the-art methods, and offer insights into future directions. We expect our survey to provide novel insight and inspiration, facilitating knowledge of deep AL and encouraging study on the open topics presented here.

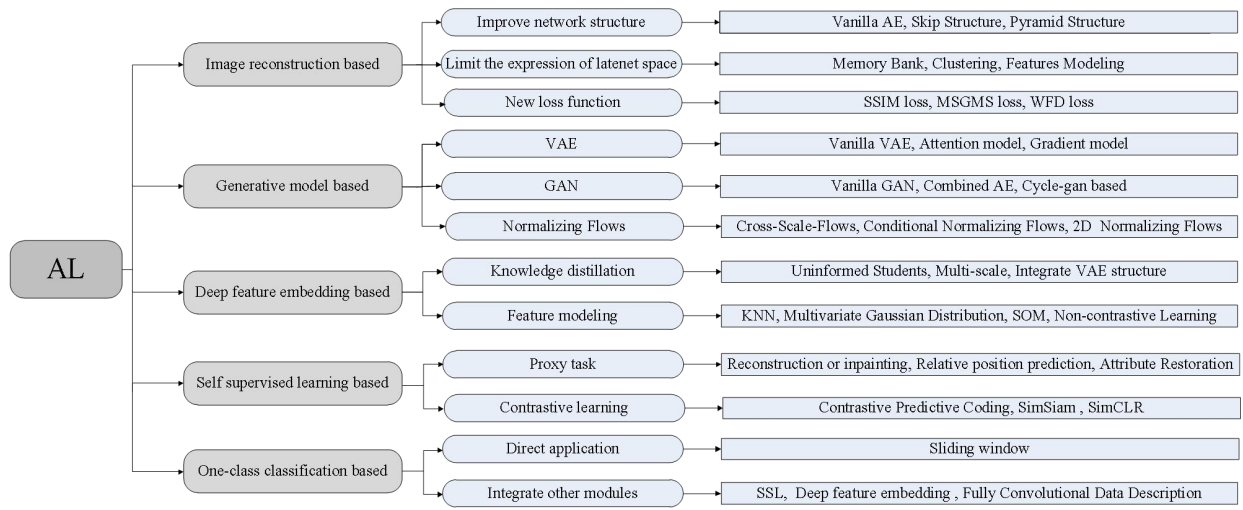

Fig. 3. A taxonomy of deep learning-based methods for unsupervised AL.

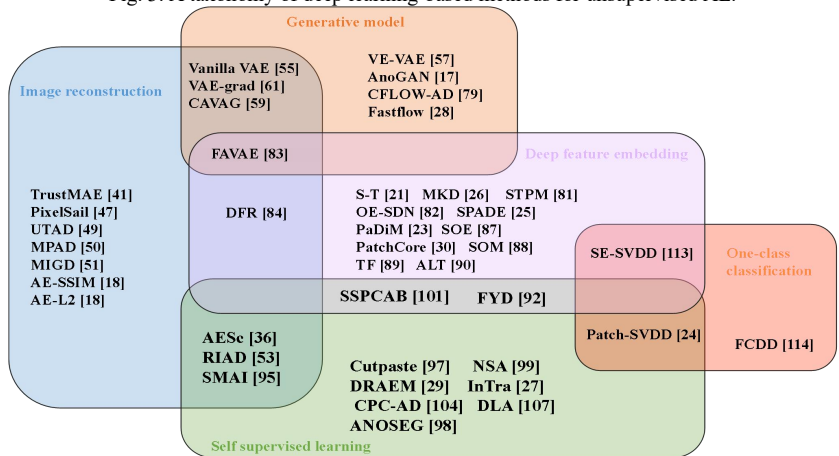

Fig. 4. Venn diagram of past major AL methods, divided into five categories. Some methods fall into more than one category, listed in the overlapping region.

## C. Contributions to this Article

This paper summarizes the surprising success and dominance of deep AL in industrial images but excludes other areas such as medical images [12] and video AL [13]. Although some of the strategies have been validated in the above scenario, real industrial images lack a priori knowledge of medical images and video sequence information. The following are the main contributions of this paper:

• To the best of our knowledge, this is the first work to focus specifically on deep learning algorithms for unsupervised AL using industrial images. At present, most of the surveys on AL or surface defect detection in industrial scenes focus on supervision methods.

• We present a taxonomy (refer to Fig. 3) that covers the latest and most advanced methods in deep learning for AL. We give a more detailed sub-classification framework than previously outlined. In addition, we draw the typical AL network under the unsupervised paradigm with a Venn diagram (refer to Fig. 4), which is convenient for readers to understand the distinction and correlation between methods.

• A comprehensive comparison of existing methods on a public dataset is provided, while we also present a summary and insightful discussion.

The rest of this paper is structured as follows: Section II summarizes the problems and related developments in AL during the previous five years. After that, Section III contains a taxonomy of existing deep learning-based approaches. The following Section IV summarizes the benchmark datasets, and presents an overview of evaluation metrics and their corresponding performance. Finally, Section V concludes this paper with an important future research outlook.

## II. BACKGROUND

### A. Problems and Challenges

The goal of AL is to find anomaly regions using only defect-free training samples. The anomalies are defined as the observations that deviate significantly from some notion of normalcy. In general, anomalies in industrial scenarios are divided into two types. (i) **textual anomalies** with little semantic information, and (ii) **functional anomalies** with a considerable amount of semantic information. To better illustrate this distinction, we use images from the MVTec AD [20]. Texture anomalies account for a large percentage of industrial defect detection, such as cracks in bottles, marks on hazelnuts, and scratches on the surface of the wood. These can be regarded as variants of local pixels on the overall texture, as shown in the first row of Fig. 5. Functional anomalies differ from textural anomalies, which often do not have textural variations but contain semantic information. In the second row of Fig. 5, a subtle anomaly is concerned with whether the needle is inserted into the hole. This anomaly necessitates high-level semantic information, making it more difficult to detect than a textural anomaly.

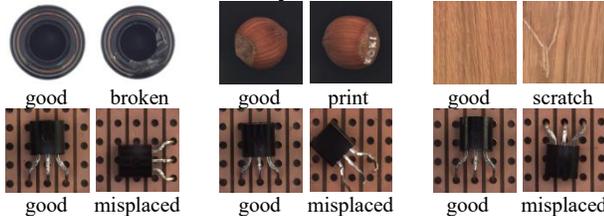

good　　broken　　　　good　　　print　　　　good　　scratch

good　misplaced　　good　misplaced　　good　misplaced

Fig. 5. Illustration of the types of anomalies: textural anomalies and functional anomalies. Low-level textural anomaly: a scratch on the wooden surface. High-level semantic anomaly: transistor with an offset pose or a pin not inserted in the hole. Note that the texture of a misplaced transistor is similar to that of a normal/good sample.

In the literature, multiple closer terminologies are used, such as image segmentation, image saliency detection, surface defect defection, and novelty detection. We here explain the difference between AL with the other terminologies. *Image segmentation* is a broad concept. To some extent, AL is equivalent to unsupervised image segmentation. But image segmentation mostly focuses on acquiring specific objects with semantic information, which may not be abnormal. *Image saliency detection* can be defined as the task of finding saliency regions, which often correspond to the important object in the image. However, some anomalies may not be salient in the whole image, such as the functional anomalies in Fig. 5. *Surface defect detection* and anomaly location are very close concepts in industrial scenes. We can simply regard the anomaly location of industrial images as quite equivalent to unsupervised pixel-level defect detection. *Novelty detection* refers to image-level classification settings, where the inlier and outlier distributions differ significantly, which is quite similar to AD.

Moreover, there are significant challenges for AL in real industrial scenarios, as follows:

**Training sample distribution problem:** All the training samples used for unsupervised AL are defect-free. The degree of balance in the distribution of defect-free samples influences the judgment of anomaly location; for example, if a particular normal sample or region is missing from the training data, the trained model may identify that normal sample or region as anomalous. In other words, the goal is to make the machine's perspective as compatible with human experience as possible. Furthermore, there is the possibility of contamination or data noise in normal data in complex industrial scenarios. Variations in imaging conditions, such as illumination, perspective, scale, shadows, blur, and so on, can result in significant differences in training samples that should not be considered anomalies.

**Multi-scale anomaly problem:** In real industrial scenes, some anomalies, such as cracks, are often subtle and occupy a tiny area. These small areas may even occupy only a few pixels in the entire high-resolution image. Thus, in anomalous images, tiny pixels are easily overwhelmed by normal conditions rather than anomalies. Furthermore, large span anomalies are also common in real-world scenes. It is, therefore, a challenge to locate anomalies by taking into account both small, subtle defects, and large defects with a complete span.

**Fine boundary problem:** The decision boundary of the model should be equal to the ideal distribution boundary. However, because of the scarcity of pixel-wise supervised labels, comprehensive segmentation of precise anomaly contours is another challenge in anomaly location. Currently, most anomaly location methods' location accuracy is insufficient, significantly different from the ground-truth.

### B. A Road Map of Anomaly Localization

AL for industrial images has a brief history, dating back to the research of [14-16]. Most non-deep-learning-based AL models rely on sparse coding [14, 15] and dictionary learning [16]. Since 2017, a growing number of deep AL methods [19] have emerged due to the rousing success of deep learning techniques in computer vision. GAN models [17, 22] and AE reconstruction networks [18] were first used in the deep AL models. To consistently compare the effects of AL, a complete industrial AL dataset was proposed by MTVec company [20]. Later, feature embedding-based models, which are more effective and efficient, became the prevalent AL architecture. Knowledge distillation [21, 26] and pre-trained feature comparison [23, 25, 30] are examples of representative models. Then several self-supervised learning-based methods have been applied to AL tasks [24, 29]. Flow-based models [28] and transformer models [27] as better approaches have also been embedded into AL networks. A brief chronology of AL is shown in Fig. 6. Despite its short history, AL research has produced hundreds of papers, and we have comprehensively selected influential papers published in prestigious journals and conferences; this survey focuses on major advances in the past five years.

### III. TAXONOMY

This section summarizes the unsupervised AL methods in terms of the high-level paradigm. Specifically, we review the various types of AL models given in Fig. 3, with subsections devoted to each category. In each subsection, we take a further breakdown of its representative works. However, some of the work fall into more than one category. Therefore, in Table 10, we divide the works according to the Venn diagram of Fig. 4, and the overlapping area includes the cross part of the methods.

### A. Image Reconstruction-based Approach

The first group is the 'image reconstruction-based methods', which are the most basic AL methods. It is based on the idea that the model is trained to reconstruct only normal images; and then when an abnormal image is input, the model still reconstructs the anomalous region as normal, *i.e.*, the model cannot reconstruct the abnormal image correctly. Therefore, the difference between the input and reconstructed images represents the localization result. As illustrated in Fig. 7, the input image is compressed on a low-dimensional bottleneck layer (latent space). This model assumes that the data has a high degree of correlation/structure. Consequently, the encoder compresses the data into an intermediate representation, which is then employed by the decoder to reconstruct the input image.

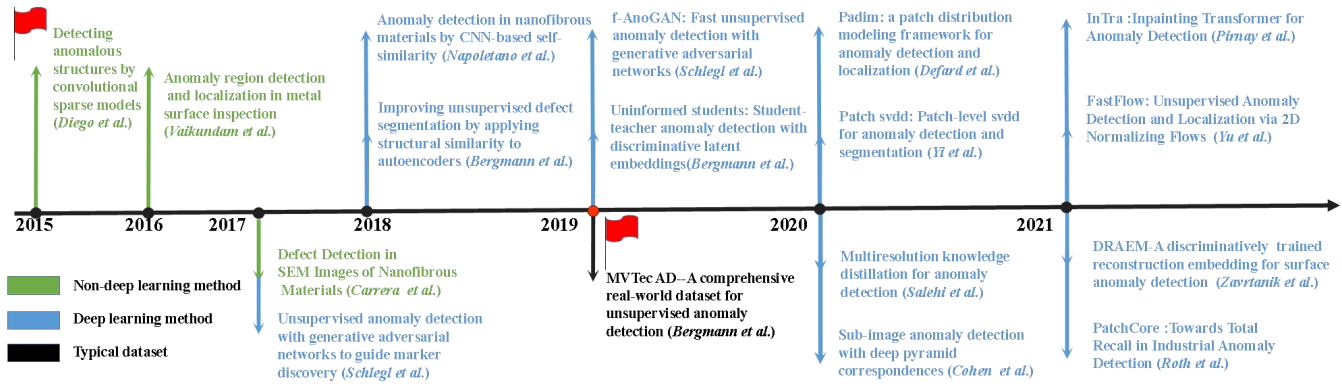

Fig. 6. A brief chronology of AL. Typical highly cited milestone methods are mentioned.

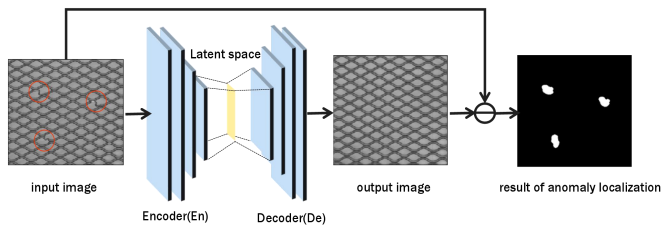

Fig. 7. Illustration of the main components of an Autoencoder (AE).

The earliest ***vanilla AE*** was used for unsupervised anomaly segmentation with brain MR images [31]. The industrial image reconstruction-based methods in AL follow this idea of the AE series. Youkachen et al. [32] used a convolutional autoencoder (CAE) for industrial image reconstruction. By sharpening the differences between the reconstructed and input images, they generated the final segmentation results of the surface defects in the hot-rolled strip. Kang et al. [33] reconstructed overlapping patches instead of the insulator image for detecting insulator surface defects, since the direct reconstruction of the entire image is intractable, and the defective area is usually a tiny part. Chow et al. [34] presented an application of deep learning in implementing AL of structural concrete defects to facilitate visual inspection of civil infrastructure. It also crops the input image and then feeds patches to vanilla AE for reconstruction. However, these ***vanilla*** AE methods may suffer from challenges due to the complex industrial scenarios. Here we summarize the novel designs of the AE-based image reconstruction framework for AL.

1) *Improvement of Network Structure:* Different from the vanilla AE, two simple structure improvements are proposed to enhance the reconstruction ability better. The first one is skipping layers. Skip-GANomaly [35] employed an encoder-decoder convolutional neural network with skip connections to thoroughly capture the multi-scale distribution of the normal data distribution in high-dimensional image space. Based on an evaluation across multiple datasets from different domains and complexity, the skip connections provide more stable training and achieve numerically superior results than vanilla AE. Collin et al. [36] proposed an autoencoder architecture with skip connections for AD in industrial vision to increase the sharpness of the reconstruction. Besides, some works extended the design of AE and feature pyramid combination to the multi-scale anomaly perception. Mei et al. [37] reconstructed image patches at different Gaussian pyramid levels with AE and synthesized the reconstructed results from these different resolution channels. Yang et al. [38] proposed a multi-scale feature clustering-based fully convolutional autoencoder (MS-FCAE) method, which utilizes multiple feature AE subnetworks at different scale levels to reconstruct several textured background images. Mishra et al. [39] focused on image AD using a deep neural network with multiple pyramid levels to fuse image features at different scales.

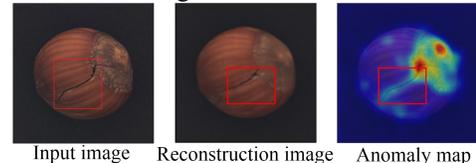

Fig. 8. Illustration of the strong generalization of AE.

However, the aforementioned improving structures do not work well on complicated textured or object datasets. Since some research shows that as AE employs a bottleneck layer to reconstruct the input image, it is difficult to manage its generalization ability. When the AE's generalization ability is powerful, anomalous features are confused with the normal feature, resulting in the network's output exactly reproducing the input. As depicted in Fig. 8, these models tend to directly copy the scratch area (marked with a red rectangle) as the output, resulting in missing anomalies. Due to the above reasons, many contemporary methods attempt to constrain latent space representation.

2) *Constraining the Representation of Latent Space:* According to how to deal with the features, we further group these methods into memory banks, clustering, and features modeling.

a) ***Memory banks*** employ a form of dictionary learning to replace the original expression of the latent space. Gong et al. [40] were the first to employ memory banks to detect anomalies. The memory bank module in this model is a matrix with each element similar to a word in dictionary learning and capable of encoding defect-free sample features. In particular, only a limited number of words are used for reconstruction during the training phase, prompting each matrix element to represent each row. Hence, the normal samples are indexed to the most comparable elements for good reconstruction, while the difference between the abnormal sample and reconstruction is magnified as the anomaly score. Later on, many following works [41, 42, 43, 44] adopted this design. Unlike prior memory bank approaches, SAP2 [45] constructed memory banks from pre-trained features for AD and localization. Liao [46] proposed a new AL framework by learning latent representations with selecting and weighting in a batch operation. This model is essentially a simplified version of a memory bank.

TABLE 2 Strengths and weaknesses of different image reconstruction-based AL approaches

| Taxonomy | Methods | Strengths | Weaknesses |
|---|---|---|---|
| Improvement of network structure | Skip layers | Using a skip layer to enhance reconstruction performance | Reconstruction easily fails in complicated textured or object datasets. |
| | Feature pyramid | Suitable for multi-scale anomalous regions | |
| Constraining the representation of latent space | Memory banks | Reduces the strong generalization ability of the AE network | Hard to determine the optimal constrain parameters and is limited by the effect of pixel-level comparison. |
| | Clustering | Aims to cluster the effective information for latent representation | |
| | Modeling features | Restrict the distribution of latent space to a specific distribution | |
| New loss function | SSIM [18] | Add the luminance, contrast, or structure information for loss | Localization effect has not been significantly improved compared with the original loss. |
| | MSGMS [53] | Introducing the image gradient information for loss | |
| | WFD [54] | Transfer the image to the frequency domain to calculate the loss | |

b) *Clustering* of latent space features is another way to enhance the discrimination of the model. Yang et al. [38] proposed a feature clustering module in MS-FCAE to enhance the discriminability of the encoded features in the latent space, which improves the reconstruction accuracy of the texture background image. An anomaly feature-editing-based adversarial network for texture defect visual inspection is proposed in [48], in which the latent space of the AE module also utilized feature clustering. Moreover, some classical clustering operations for latent space have been proposed, including the standard K-means clustering [50].

c) *Modeling features* of the latent space is also an effective way of limiting the representation. In [47], a discrete latent space probability model was estimated using a deep auto-regressive model, named PixelSail. It determines the latent input space regions that deviate from the normal distribution during the detection stage. In particular, the deviation code is then resampled from the normal distribution and decoded to provide a restored image closest to the anomalous input. The anomaly region is identified by comparing the restored and anomaly images. In addition, some approaches to modeling latent space features have also been proposed, including the Gaussian descriptors [51] and even graph network models [52].

As the image reconstruction-based methods usually employ a pixel-level comparison metric, AE networks choose trained loss with $L_1$-distance and $L_2$-distance. This results in the comparison of inputs and outputs being only at the pixel-level and lacking semantic information. Therefore, some improvements based on the loss function have been proposed. For this kind of method, the key problem is considering the semantic information in the image reconstruction effect. We discuss this problem in the following.

3) *New Loss Function:* Bergmann et al. [18] were the first to use the structural similarity (SSIM) metric in image reconstruction. In contrast to pixel-wise comparisons, SSIM loss considers a region's brightness, contrast, and structural information. Compared to $L_2$ loss, the SSIM loss significantly improves the performance of AL in the textured datasets. In [53], a new multi-scale gradient magnitude similarity (MSGMS) loss was proposed, which pays more attention to the structural differences in the reconstruction. The MSGMS loss is constructed by calculating the gradient images of the original and reconstructed images. The overall AUROC on the MVTec AD improved by 6.5% when using MSGMS. Nakanishi et al. [54] designed a new loss function named weighted frequency domain (WFD) loss, which transforms the reconstruction loss calculation from the image domain into the frequency domain. It provides a sharper reconstructed image, improving anomaly location accuracy.

*Brief Summary:* Table 2 gives a glance at these three types of image reconstruction-based methods, and analyzes their advantages and disadvantages. Although image reconstruction-based methods are usually very intuitive and interpretable, their performance is limited because AE does not introduce any prior knowledge, and its effect only depends on the expression ability of latent layer to defect-free features.

### B. Generative Model-based Approach

In order to overcome the shortcomings of AE-based methods with poor reconstruction performance, generative models are introduced into the industrial AL field. The basic idea behind generative models is to model the real data distribution from the training data and then utilize the learned model and distribution to generate or model new data. The key to AL in this framework is explicitly or implicitly obtaining the feature distribution of defect-free data. As the generative model only generate normal samples, the difference between the generated or reconstructed samples and the input is the abnormal region. Unlike AE, which only considers the final reconstruction, the generative model could reflect this difference in latent or feature space. According to different models, we further group these methods into VAE (Variational Auto-Encoder), GAN (Generative Adversarial Network), and NF (Normalizing Flow).

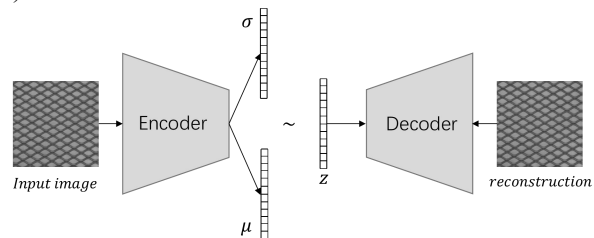

Fig. 9. Graphical illustration of VAE.

1) *Variational Auto-Encoder (VAE):* As depicted in Fig. 9, VAE introduces a prior distribution for normal samples in the latent space, which is typically a multidimensional standard normal distribution. This indicates that the encoder output is no longer simply latent space but rather an estimated distribution, and therefore this approach is a subset of modeling the latent space features in the AE-based methods. Thus, the difference between vanilla VAE and AE is the additional loss employed to assess the difference between the estimated distribution and the prior distribution, such as Kullback-Leibler (KL) divergence loss. Matsubara et al. [55] first introduced VAE for industrial AL, which was evaluated on a toy dataset and real-world manufacturing datasets. Kozamernik et al. [56] proposed a VAE-based model for visual quality control of KTL coatings. By calculating the negative log-likelihood of the distribution returned by the decoder, anomalies containing surface defects are successfully detected. Although the vanilla VAE successfully located the anomalies, the localized accuracy for anomaly regions in [55, 56] is relatively poor. Some researchers have tried to add other mechanisms to VAE for more fine-grained AL.

a) *Attention-based methods*: Liu et al. [57] first proposed a technique to generate VAE visual attention using gradient-based attention computation. The generation method of attention map is similar to grad-CAM [58]. In particular, the corresponding weight coefficient is obtained based on computing gradients of the latent space variable with respect to the last layer feature maps of the encoder. The final attention map is then generated by weighting the feature map of the last layer of the encoder. The apparent region in the acquired attention map is the anomalous one when detecting anomalous images. Venkataramanan et al. [59] proposed a convolutional adversarial VAE with guided attention (CAVGA), which localizes the anomaly within a latent convolutional variable to preserve the spatial information. It generates an attention map following the main idea of [57], with the expectation that the attention map generated by the training network could cover the entire image.

b) *Gradient-based methods*: According to Zimmerer et al. [60], the loss gradient with respect to input image gives the direction towards normal data samples, and its magnitude could indicate how abnormal a sample is. Benefiting from this conduction, Dehaene et al. [61] proposed the gradient descent-based VAE. As seen from the reconstructed images in [61], the gradient descent-based method gives better quality reconstructions than vanilla VAE. In [123], Chu et al. proposed that the change in loss values during training can also be used as a feature to identify anomalous data. The algorithm is thoroughly evaluated and compared against other baselines on two datasets, MVTec AD [115] and NanoTWICE [16], which spans a large variety of different objects and textures.

2) *Generative Adversarial Network (GAN):* GAN-based models are classified into three types based on their network structure, as below.

a) *Vanilla GAN*: Schlegl et al. [17] were the first to apply GAN to localize anomalies. The generative network $G$ in this approach receives randomly sampled samples from the latent space as input, and its output must be as close to the real samples in the training set as possible. The discriminative network $D$ takes input from either the real samples or the output of the generative network, and its goal is to differentiate the output of the generating network from the real samples as much as possible. The whole loss includes two-part: the reconstruction loss of the $G$ and the feature difference loss of the $D$. The difference between the output of the generating network $G$ and the input image determines anomalous regions. Later on, several following works [62, 63] adopted this model for the industrial surface defect.

b) *GAN combined with AE*: As the vanilla GAN employs a single image as input in the inference stage, the network must frequently repeat to find the optimal latent space vector to achieve the desired generation result. Some joint AE structured GAN methods have been proposed in response to the drawback that vanilla GAN needs to update its parameters constantly.

i) *Improving the input of generator G* is the most straightforward way to train a GAN-based AL network where the input is changed to a real defect-free image instead of the randomly sampled samples from the latent space, and therefore the generative network $G$ is accordingly changed to a complete encoding-decoding structure, as shown in Fig. 10 (A1). This improvement is equivalent to employing a discriminator $D$ on the image reconstruction method to distinguish whether the image is a real input defect-free sample or a reconstruction sample. This GAN-based AL approach was used by Balzategui et al. [64] to implement the quality inspection of monocrystalline solar cells and Hou et al. [43] to form a divide-and assemble the framework for AL. In addition, some methods used defective synthetic samples as input to the generator G, as shown in Fig. 10 (A2). This indicates that the generator is implementing repair or inpaint in this case. Zhao et al. [65] established the network by repairing defect areas in the samples, and then compared the input sample and the restored one to indicate the accurate anomaly regions. In particular, a denoising autoencoder generative adversarial network is proposed by Komoto et al. [66], detecting defective regions by recovering a defective product image that adds an artificial defect to a defect-free product image.

ii) *Improving the generator G* is another common approach that aims to employ constraints on reconstructing latent space features. Akcay et al. [67] proposed the GANomaly network, which is an additional encoder after the auto-encoder, forming an "encode-decode-encode" structure, as shown in Fig. 10(B1). The difference between the second encoder's output and the first encoder's output is employed to assess whether the input is anomalous. This similar structure is also followed by cigarette packet anomaly location [68], industrial surface detection [69], and textured surface detection [70]. Besides, Schlegl et al. proposed another scheme, namely f-AnoGAN [22], which fixes the trained decoder in the generator $G$ and reuses it as a generator for the latent space reconstruction network. It is worth noting that the strategy of constructing the reconstruction training of feature vectors in latent space is embraced, as depicted in Fig. 10(B2).

iii) *Improving the discriminator D* is generally made by employing multiple discriminators to enhance the discriminative ability of the GAN network. As shown in Fig. 10(C1), Zhang et al. [71] proposed that DefGan design an additional branch of the reconstructed image through a latent space pitting operation and a weight sharing, which form a new discriminative loss together with the original input image. Li et al. [72] designed an additional latent-space discriminator by constructing a new latent-space feature through random sampling, which was fed into the designed discriminator with the latent-space feature from the original generator for discrimination, as shown in Fig. 10(C2). A summary of past representational work, covering the structure, year, and description of the imported GAN-based AL model, is presented in Table 3.

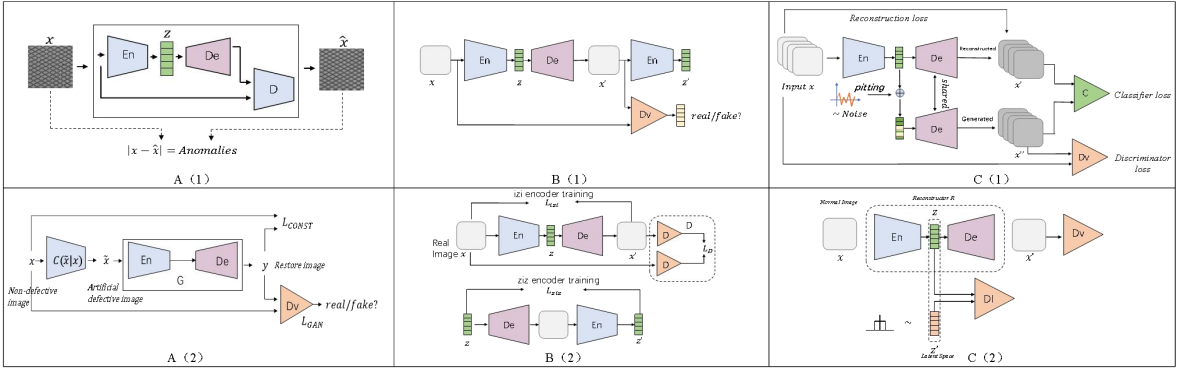

Fig. 10. Pipelines of improved GAN-based AD networks: (A1-A2) improving the input of the generator $G$, (B1-B2) improving the generator $G$, and (C1-C2) improving the discriminator $D$.

TABLE 3 Representative works of improved GAN-based AL networks

| Approach | Representative networks | Year | Description |
|---|---|---|---|
| improving the input | DAE-GAN [66] | 2018 | The input is either a defect-free image or an artificially defective image. The generator is an AE structure for image reconstruction or inpaint |
| improving the generator $G$ | GANomaly [67], f-AnoGAN [22] | 2018 | Reconstruction of latent space variables. Generators are "encode-decode-encode" structures or multiplexed "encode-decode" structure |
| improving the discriminator $D$ | DefGAN [71] | 2020 | Introduction of multiple discriminators to improve the generation of normal samples |

TABLE 4 Representative works of normalizing flows for AL

| Method | Venue | Year | Description |
|---|---|---|---|
| DifferNet [78] | WACV, 2021 | Aug. 2020 | Image transform for multi-scale evaluation, vanilla normalizing flow block, AL by back-propagating the negative log-likelihood loss to the input image |
| CFLOW-AD [79] | WACV, 2022 | Jul. 2021 | Multi-scale pyramid pooling, conditional vectors concatenating with the decoder coupling layers in NF block |
| CS-Flow [80] | WACV, 2022 | Oct. 2021 | Multi-scale feature maps jointly, a fully convolutional NF with cross-connections between scales |
| Fastflow [28] | Arxiv, 2021 | Nov. 2021 | Vision transformer (ViT) as the feature extractor, 2D Flow Model |

c) *CycleGAN:* Thanks to the developed GAN technology, the leverage of muti-GAN to establish mappings between different feature domains has become easier to implement. The CycleGAN framework consists of four CNNs, namely two generators and two discriminators. While the generators try to learn the mapping between the respective domains, the discriminators try to discern between real and synthesized images within one image domain. Generally, the CycleGAN-based approach has two different domains. Yu et al. [73] proposed an adversarial image-to-frequency transform (AIFT) network applied in unsupervised AL of road cracks. Another transformation between the image domain of a defect-free and the image domain of a synthetic defect is a classical approach, which is also the primary way of generating defect samples in defect detection [74]. Some works, e.g., [75] and [76], applied CycleGAN to rail defect detection and fiber anomalies inspection. However, the CycleGAN-based methods were validated on specific datasets, and lack severe results on commonly used publicly available datasets (e.g., MVTec AD [115]). Therefore, their effectiveness needs further validation. Also, it may be noted that the lack of data samples poses a challenge in training two GAN networks well at the same time in industrial scenarios.

3) *Normalizing Flow (NF):* Different from previously introduced generated models that cannot estimate accurate data likelihoods, normalizing flows (NFs) [77] are neural networks that learn transformations between data distributions and well-defined densities [78]. The forward pass projects data into a latent space to calculate exact likelihoods for the data given the predefined latent distribution. Conversely, data sampled from the predefined distribution can be mapped back into the original space to generate data. For the AL task, the anomaly region is obtained by measuring the distance between the feature of the test image and the estimated distribution of defective-free images. Instead of dealing with the image directly in the VAE or GAN-based methods, NF-based methods perform AL on features. Most currently available NF-based methods first leverage pre-trained networks to extract normal image features and then employ NF models to estimate the corresponding distributions accurately. The first one is **DifferNet**, proposed by Rudolph et al. [78]. This model utilizes a normalizing-flow-based density estimation of image features at multiple scales. In particular, the AL result is generated by back-propagating the negative log-likelihood loss to the input image, similar to grad-CAM. However, this framework focuses on image-level anomaly classification, which is not optimized for the localization of defects on images. The anomalous localization areas in MVTec AD do not accurately fit the ground-truth range. In 2021, three NF-based AL methods were improved in three different ways and achieved surprising results on multiple datasets. Gudovskiy et al. [79] designed the CFLOW-AD model based on a **conditional normalizing flow** framework for AL. In particular, a conditional vector using a 2D form of conventional positional encoding (PE) is proposed, then concatenating the intermediate vectors inside decoder coupling layers with the conditional vectors. CFLOW-AD achieves new state-of-the-art for famous MVTec AD with 98.62% AUROC and 94.60% AUPRO in localization. To boost the fine-grained representations incorporating the global and local image context, Rudolph et al. [80] proposed a fully convolutional ***cross-scale normalizing flow*** (CS-Flow) that jointly processes multiple feature maps of different scales. The convolutions in the CS-Flow block were performed at two levels, with cross-connections between scales at the second level. However, many non-anomalous backgrounds still appear in the final localization results. Recently, Yu et al. [28] proposed a new AL network called Fastflow, which is similar in detection principle to the previous works, except that it designs a ***2D flow*** based on "$3 \times 3$" and "$1 \times 1$" convolutions. It firstly utilized the visual transformer as a feature extractor for normal samples, and the features are then fed into a post-stage flow model for estimating the probability distribution. This model achieves excellent results in the MVTec AD with 98.5% pixel-AUROC. In summary, a comparison of four typical flow-based AL methods is shown in Table 4.

TABLE 5 Strengths and weaknesses of different generative model-based AL approaches

| Taxonomy | Methods | Strengths | Weaknesses |
|---|---|---|---|
| VAE | Attention-based | Anomaly map is generated by derivation, not reconstruction | Localization result of the anomalous region is coarse. |
| | Gradient-based | Changing in loss values during training can be used as a feature to identify anomalous data | |
| GAN | Improving the input | Employs GAN and its modifications for enhancing the ability of image reconstruction or generation | Extensive training costs. Generator might become unstable. The generation effect of normal areas in the image is poor, which leads to false detection. |
| | Improving the generator G | | |
| | Improving the discriminator D | | |
| NF | CFLOW-AD [79] | Can estimate accurate data likelihoods for normal samples | The model needs fine design, and the design criteria are different from vanilla CNN. |
| | CS-Flow [80] | | |
| | Fastflow [28] | | |

***Brief Summary:*** Table 5 provides a concise overview of these three types of generative model-based approaches, as well as a brief discussion of their pros and cons. The generation effect of VAE or GAN over normal areas in the image is poor, which easily leads to false detection. At present, the best localization result is achieved by NF, which combines the deep feature embedding-based methods discussed in the following subsection.

### C. Deep Feature Embedding-based Approach

Although image reconstruction or generative models succeed in several industrial scenes, several works observed that this method typically produces incorrect reconstruction results due to the lack of feature-level discriminatory information. As depicted in Fig. 11, the reconstruction part ignores the details of the image (marked with a green rectangular box). The defective-free area at the hazelnut base is not well reconstructed, which leads to over-detection problems.

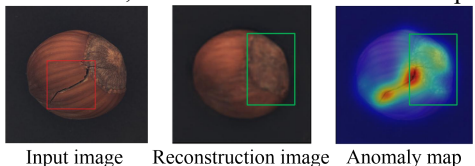

Input image    Reconstruction image    Anomaly map
Fig. 11. Illustration of the poor image reconstructions.

To overcome the limitation of image reconstruction or generative models, another line of research proposes to employ deep feature embedding-based methods, which are generally divided into two parts: feature extraction and anomaly estimation. Hence, the final pixel-level anomaly map is generated by comparing deep embedding features from target and normal images. Feature extraction part usually chooses pre-trained on large-scale databases such as ImageNet, or self-supervised learning. In particular, the NF mentioned above can also be regarded as the deep feature embedding-based method combined with the generative model. According to the paradigm of anomaly estimation, we further group these methods into 'knowledge distillation-based' and 'deep feature modeling-based'.

1) *Knowledge Distillation-based Methods:* In order to better embed the deep feature information, a student-teacher framework is leveraged here. The teacher model plays as a pre-trained feature extractor, and the student model is used to estimate a scoring function for AL. Bergmann et al. [21] proposed the ***uninformed students***, which is the first to employ the knowledge distillation model for anomaly location. It is characterized by employing one teacher and multiple students. Student networks are then trained to regress the output of a descriptive teacher network that was pre-trained on a large dataset of patches from natural images. In particular, anomalies are localized when outputs of the student networks differ from those of the teacher network and differences in the output for different student networks. However, this model only employs the output of the last layer of the network as a feature for knowledge distillation, and the multi-patch approach is adopted to localize the anomaly better, which puts a burden on computing time. To deal with the above limitation, Salehi et al. [26] presented a ***multi-resolution knowledge distillation*** approach, where considering features from multiple intermediate layers in the distillation process leads to better use of the expert's knowledge and more significant differences compared to solely utilizing the output of the last layer. In particular, its AL map is generated by back-propagating the loss to the input, which also leads to limitations in its localization effectiveness. Wang et al. [81] have further extended the technique of the multi-scale AL method by introducing the student-teacher feature pyramid matching (STPM) model. Their AL map is generated by directly calculating the differences between the multi-feature layers of the teacher network and the student network. This model enables accurate localization results and avoids the input image's path-size setting. On the MVTec AD, it achieved 98.5% AUROC and 92.1% PRO score. Besides, some works also extend the knowledge distillation framework into AE or VAE for better reconstruction results. Chung et al. [82] evaluated an outlier-exposed style distillation network (OE-SDN) that mimics the mild distortions caused by an AE, termed style translation. This approach utilizes the difference between the outputs of the OE-SDN and AE as an alternative anomaly score. Dehaene et al. [83] proposed a feature-augmented VAE (FAVAE) architecture consisting of a feature extraction module with VAE architecture, where the output of the extraction module is correlated with the multilayer output of the decoder in the VAE. It can be regarded as the knowledge distillation operation.

*2) Deep Feature Modeling-based Methods:* In this pipeline, a feature space is first needed to build for the input image, then to realize the measurement or comparisons of the features by feature modeling. These tricks can be clustering, or some probability distribution fitting, or some learning models. Compared with the knowledge distillation method, it often employs one end-to-end network with no distinction between teacher and student networks. Cohen et al. [25] presented an alignment-based method for detecting and segmenting anomalies inside images. It constructs a pyramid of features using a pre-trained Wide-ResNet50 model and employs these feature maps to find the *K* nearest anomaly-free images. Defard et al. [23] designed a patch distribution modeling (PaDiM) framework that first generated features using a pre-trained CNN, then modeled normality by applying a multivariate Gaussian distribution to each location. In the test stage, the final anomaly map is generated by measuring the Mahalanobis distance of the features at each location to a "standard feature template". Since this method was the best reported result until the advent of NF, some following works [85, 87, 88, 89] adopted this design. As the method models the fixed positions of feature maps, it is only adapted to aligned datasets. In [84], Yang et al. proposed AE-based feature reconstruction to replace the previous Gaussian distribution modeling strategy. However, the results of AL did not outperform PaDiM. It does not perform well on the typical datasets Tile and Wood of MVTec AD. Mishra et al. [85] presented VT-ADL, which combines the traditional reconstruction-based methods with the benefits of Vision Transformer (ViT). The input image is encoded using a ViT, and its resulting features are then fed into a decoder to reconstruct the original image. Moreover, a Gaussian mixture density network models the distribution of the transformer-encoded features to estimate the distribution of the normal data in this latent space. Its structure is complex but demonstrates that the visualization of the anomaly map on MVTec AD is poor. In [86], multilayer feature sparse coding (MLF-SC) is employed for AD. The AL results still rely on the image-pixel level reconstruction effect. Kim et al. [87] followed the idea of the PaDiM and devised a random feature selection method extending to semi-orthogonal embeddings (SOE) to avoid the computational complexity of the multidimensional covariance tensor. It achieved good results for MVTec AD, KolektorSDD [117], and KolektorSDD2 [118]. Roth et al. [30] followed the idea of the SPADE and proposed PatchCore for AL. It employed the nominal patch-level feature representations extracted from ImageNet pre-trained networks and minimal runtime through coreset subsampling to realize a low computational cost. On MVTec AD, this method achieved more than 99% of image AD AUROC but attained inaccurate localization results. Li et al. [88] also followed the idea of PaDiM and used a self-organizing map (SOM) rather than a multidimensional Gaussian. This model has a slightly better AUROC at pixel level on MVTec AD than the original PaDiM. Rippel et al. [89] introduced fine-tuning the learned representation in the feature exacted part, thus improving the original PaDiM for AL. Yan et al. [90] proposed a multi-level image reconstruction and feature comparison approach to AL with an adaptive attention-to-level transformation (ALT) strategy. ALT simultaneously adjusts the weights of reconstruction levels and feature measurement scales to utilize consistent levels of features for reconstruction and AD.

TABLE 6 Comparison of past works with some key elements

| Methods | Year | Pre-trained | Normal images usage | Anomaly map | Multi-Scale | Pros | Cons |
|---|---|---|---|---|---|---|---|
| SPADE [25] | 2020.5 | WideResNet50 | KNN | Euclidean distance | DNN scales | Simple structure, no training required | The complexity of the KNN algorithm operations is linearly related to the training samples. The more training images there are, the greater the storage requirement |
| PaDiM [23] | 2020.11 | WideResNet50 | Multivariate Gaussian distribution modeling | Mahalanobis distance by fixed positions | DNN scales | Distributed estimation improves anomaly location performance | It uses separate estimation distributions at fixed locations, with significant performance degradation when the dataset is unaligned |
| DFR [84] | 2020.12 | VGG19 | AE-based feature reconstruction | Reconstruction error | DNN scales | Simple learning network | Incomplete localization area. Does not perform well in many data sets, e.g., tile, cable, and transistor |
| VT-ADL [85] | 2021.4 | ViT | Gaussian mixture density network | Reconstruction error | ViT | An early introduction of ViT feature coding | The structure is complex, and the overall performance is not good |
| MLF-SC [86] | 2021.4 | VGG16 | Sparse coding | Top-5 largest reconstruction errors | DNN scales | Combining sparse coding methods | Poor reconstruction result |
| Semi-orthogonal embedding [87] | 2021.5 | WideResNet50 | Random feature selection, into semi-orthogonal embedding | Mahalanobis distance | DNN scales | Retaining the better performance by avoiding redundant sampling | The improvement over PaDiM is not significant and does not outperform PaDiM on tile and wood |
| PatchCore [30] | 2021.6 | WideResNet50 | KNN with the memory bank | Patch distance | Patch | Faster inference and reduced feature storage capacity compared to SPADE | An improved version of SPADE, but the visualization of the anomaly maps is mediocre |
| SOMAD [88] | 2021.7 | WideResNet50 | Memorizing normality via SOM | Mahalanobis distance | Patch | Self-organizing approach enhances the expression of features | Most of the anomalous regions are not refined |
| Gaussian fine-tune [89] | 2021.8 | EfficientNet-B4 | Gaussian distribution | Mahalanobis distance | DNN scales | Fine-tuning of pre-trained feature representations | Visualization results are not available |
| MLIR [90] | 2021.9 | VGG19 | Image reconstruction and feature comparison | Reconstruction error | DNN scales | Introduced weighting adjustment for different feature layers | The visualization of the AL results is similar to DFR |
| Contrario method [91] | 2021.10 | Resnet | Number of False Alarms computation | Mahalanobis distance | DNN scales | Applying statistical analysis to feature maps | The structure is complex, and the overall performance is not fine-grained |
| Focus Your Distribution [92] | 2021.10 | WideResNet50 | Pixel-wise non-contrastive Learning | Mahalanobis distance | DNN scales | Consideration of image and feature alignment | Some of the training images are inherently unalignable, and their AD regions are not fine-grained and with a lot of interference |

TABLE 7 Strengths and weaknesses of different deep feature embedding-based AL approaches

| Taxonomy | Strengths | Weaknesses |
|---|---|---|
| Knowledge distillation | AL problem is transformed into a direct feature comparison between different networks | Easy to be disturbed by the choosing layer for knowledge distillation. |
| Deep feature modeling | Rich semantic information is introduced through the pre-trained model | Memory requirements are relatively high, and the feature modeling needs careful design, which greatly impacts localization results. |

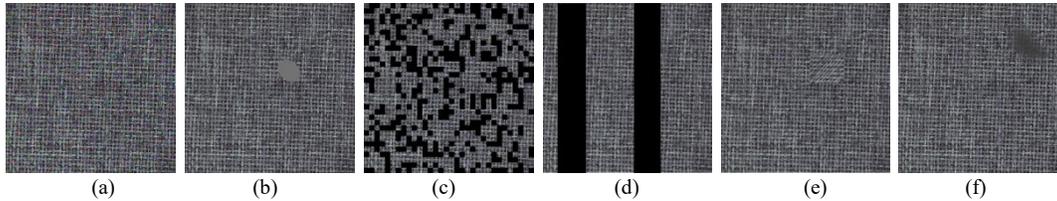

Fig. 12. Different synthetic defect images: (a) Noise, (b) Random erasure, (c) Random region mask, (d) Multi-scale striped masks, (e) Cut and paste, and (f) Composite synthesis.

TABLE 8 Strengths and weaknesses of different self-supervised learning-based AL approaches

| Taxonomy | Methods | Strengths | Weaknesses |
|---|---|---|---|
| Proxy tasks | image inpainting | Synthetic or simulated abnormal data | A gap between the simulated anomaly and the real anomaly |
| | relative position prediction | Consider the spatial information of the neighborhood patch | Correlation between neighborhood patches does not exist in many images, such as complex objects |
| | attribute restoration | Using image attributes, such as color and orientation | Effect of attribute and direction on anomaly location is limited |
| Contrast learning | CPC [105] | Using similarity to distinguish between normal and abnormal | Localization result is easily affected by interference, such as variations in imaging conditions |
| | SimSiam [106] | | |
| | SimCLR [108] | | |

Tailanian et al. [91] designed a contrario framework to detect anomalies in images, which computed the number of false feature maps before generating the final anomaly map. However, this method only achieved AUROC of 0.77 and 0.86 on Tile and Wood of MVTec's texture dataset. Recently, Zheng et al. [92] revisited the issue of unaligned data in PaDiM. They proposed the "focus your distribution" (FYD) model, which employed a coarse to fine process. Before extracting features, an image-level coarse alignment module was designed, which allowed the input image to be forcibly aligned. Pixel-wise non-contrastive learning was then utilized in the exact alignment stage, which achieved the fine alignment of dense features. This method achieved 98.2% AUROC on the MVTec AD. However, it can be seen from the AL heatmaps that there are some interference and coarse defect regions in the final result. In summary, a comparison between key elements of different state-of-the-art works is presented in Table 6. This table shows that most methods chose the Wide-ResNet50 as a pre-trained model and generated the final anomaly map by the Mahalanobis distance. Besides benefiting from the pre-trained feature, the deep feature modeling-based model may have more significant potential.

***Brief Summary***: Table 7 briefly summarizes the merits and disadvantages of several deep feature embedding-based methods to AL. Table 6 also includes the benefits and drawbacks of the representative's deep feature modeling. These models are attempting to address two following issues. The first one is to generate fine-grained and noise-resistant localization results. The second one is to extend the model to tackle multi-scale anomalies and non-aligned datasets. We believe it will garner increased interest from both academics and industry.

### D. Self-Supervised Learning-based Approach

Self-supervised learning (SSL) is the process of learning visual features from unlabeled images and then applying them to the relevant visual task. There are two SSL-based AL approaches **proxy tasks** and **contrast learning**. Proxy tasks relative focus on the development of the pretext task. On the other hand, contrast learning is primarily concerned with network design.

*1) Proxy Tasks:* The pretext task often takes many different forms, but it all boils down to predicting or recovering hidden regions or properties in an input image. Recent SSL-based AL methods have relied on three primary proxy tasks: image inpainting, relative position prediction, and attribute restoration.

a) ***Image inpainting*** is the most common proxy task. Self-supervision based on image inpainting is the same as previous methods based on image reconstruction or generation, only referred to differently. By repairing the defective synthetic images, the network model is given the ability to reconstruct normal sample images and repair abnormal regions. This kind of method can repair similar abnormal regions in the test stage. A common way of synthesizing defective images is shown in Fig. 12. The earliest defect images are generated by adding random noise; for instance, Nakazawa et al. [93] used synthesized noisy images for AD in wafer images, and the multi-scale AE method proposed by Mei et al. [37] also adopted the similar synthesized noisy images. The network model used in this approach is also known as a "denoising encoder". Besides, some data augmentation methods have been applied to generate defective training samples to improve the network's repair capability. Tayeh et al. [94] and Li et al. [95], for example, randomly erase regions in normal samples with arbitrary shapes and then fill them with a fixed color, as shown in Fig. 12(b). However, this design does not consider the structural information present in the image that facilitates subsequent network restoration. Therefore, Zavrtanik et al. [53] designed a mesh-like random mask, as shown in Fig. 12(c). The number and scale of mask regions are parameterized. On the MVTec AD, the method achieved 94.2% of the pixel AUROC. Yan et al. [96] presented a multi-scale strip mask for modeling large span defects of different scale sizes, as shown in Fig. 12(d). Some recent works have attempted to generate realistic defective images rather than just using meaningless black and white block images. Li et al. [97] were the first to crop the region on the original defect-free image and then paste it onto the image at a random angle to form a new anomaly image, as shown in Fig. 12(e). Song et al. [98] also follow this idea. In particular, some methods leverage a more sophisticated approach to background fusion, where defects are simulated by selecting different background images, varying size, brightness, and shape; and then adding image fusion to produce a more realistic defective image, as shown in Fig. 12(f). For example, Schlüter et al. [99] used Poisson fusion, Zavrtanik et al. [29] selected various texture images as defective backgrounds, and Haselmann et al. [100] borrowed from sample synthesis methods in data augmentation. Theoretically, the closer to the effect of a real synthetic defect, the more generalizable the image reconstruction and restoration capability should be. However, in real scenes, the type and shape of defects are often unpredictable, so it is difficult to decide which of the synthesis methods is optimal. As shown in Table 10, there is no relationship between the more realistic defect synthesis method and good positioning results. In general, these methods often need to be combined with designing a suitable restoration network to achieve better results.

b) ***Relative location prediction:*** Different from previously introduced models, which only consider the mapping between input and output, there is another way to assess the spatial information of the neighborhood patch. The most representative method is PatchSVDD [24], which introduces a self-supervised approach to feature extraction. It first divides the image into 3×3 patch regions, and the eight blocks around the central image block are sorted in order. Then, the encoder of this model is trained to extract informative features so that the following classifier can correctly predict the relative positions of the patches. However, due to the patch region setting, the AL results for this type of method are often very rough and not fine-grained. Pirnay et al. [27] designed an InTra network based on image restoration. Specifically, it aims at a patch region centered on $w \times w$ that can be restored by the image information of its surrounding patches, so it also leverages neighborhood information. Ristea et al. [101] designed a self-supervised predictive convolutional attentive block (SSPCAB) for AL. For each location where the dilated convolutional filter is applied, the block learns to reconstruct the masked area using contextual information. Note that this approach is essential to employ the regional features of the dilated convolution to model a more extensive range of neighborhoods.

c) ***Attribute restoration*** is characterized by using hidden attributes in the image rather than masked areas. These attributes generally include color and orientation. Fei et al. [102] proposed an attribute restoration network that turns the traditional reconstruction task into a restoration task. It first changes specific attributes of the input (e.g., removes color, changes orientation, etc.) before feeding the image into AE for reconstruction. Ulutas et al. [103] presented a split-brain convolutional autoencoder approach to detect and localize defects. Two disjointed convolutional autoencoder networks are employed to predict the sub-channel of the image from another sub-channel. Each encoder implements a conversion between different color channels. This design utilizes the color property and boosts the localization accuracy of the anomaly image.

*2) Contrast Learning:* As described before, the proxy task focuses on generating images similar to the training data at the pixel level. Another improvement is learning common features between similar instances and distinguishing differences between non-similar instances. Haan et al. [104] directly applied contrastive predictive coding (CPC) [105] to detect and segment anomalies in images. It splits the image into patches, interpreting each line of patches as a separate time step. In the test stage, the test image block is compared with a randomly selected image block in a defect-free image to calculate the contrast loss function, namely InfoNCE. The current image block is judged as an abnormal region when a certain threshold is exceeded. As a result, this method affects the detection efficiency due to the patch-based operation, and its localization accuracy is not high. In [92], the fine alignment part in the proposed network for AL is designed based on SimSiam [106]. It inputs the results of two random transformations of the same feature, extracts the features employing the same encoder $f$, and transforms them to a higher dimensional space. The predictor $g$ is employed, which transforms the result of one of the branches and matches it with the result of the other branch. This approach takes full advantage of the Siamese network's natural modeling invariance. Gui [45] followed the same idea of Siamese architecture for AL, except that the original predictor was replaced with a self-supervised module. Yoa et al. [107] presented an AL method based on SimCLR [108] (Simple framework for Contrastive Learning of visual Representations). By generating a pair of negative images in the training dataset, the design model contrasts a normal sample to a locally augmented image. This model achieved 93.4% pixel-AUROC on the MVTec AD. Self-supervised learning frameworks are still a hot topic of research, and we believe these novel models can show and validate AL, which will constitute a relevant future direction.

*Brief Summary*: Table 8 summarizes the two types of self-supervised learning-based approaches, and their benefits and drawbacks. Self-supervised learning frameworks are still a hot issue of research in general. We believe these novel models show and validate the potential of the AL, and constitute a relevant future direction.

### E. One-Class Classification-based Approach

The one-class classification approaches are usually employed for image-level AD, typical including OCSVM [109] and deep SVDD [110]. Deep SVDD trains a network and then maps the training data to a small hypersphere in the feature space. The data outside the hypersphere are called anomalies. For AL, a one-class classification-based approach locates anomalous regions by dividing the image into patches and classifying the patches into abnormal or normal categories, which enables coarse results. This form was adopted by Liu et al. [111] and Wang et al. [112] to localize anomalous regions on the steel surfaces and wind turbine blades, respectively. Furthermore, several improved versions of deep SVDD have been proposed for AL. Compared to deep SVDD, Patch SVDD [24] inspect every patch to localize a defect, and self-supervised learning is employed, allowing the features to form multi-modal clusters, thereby enhancing AD capability. In addition, the deep SVDD method was embedded in the pre-trained feature comparison by Hu et al. [113]. It estimates the pixel-wise anomalies efficiently based on deep SVDD. Liznerski et al. [114] presented a fully convolutional data description (FCDD), a modification of deep SVDD so that the transformed samples are themselves an image corresponding to a downsampled anomaly heatmap. While this approach produces a full resolution anomaly heat map, the extent of the anomaly region is not accurate due to the up-sampling operation of a fixed Gaussian kernel.

*Brief Summary*: Major AD approaches can be employed for pixel-level AL, since we can segment the complete image into several patches and then perform AD on image patches. The AD algorithm concentrates on the entire image's semantic information; therefore, the semantic information of subtle abnormal regions may be ignored. Here we observe that combining AD with some self-supervised strategies or pre-trained deep feature embedding methods is a promising way to increase localization performance.

## IV. EXPERIMENTS

### A. Datasets used by Recent Works

Five datasets are available for unsupervised learning-based AL datasets, which differ significantly in terms of image quantity, quality, resolution, and texture information.

**NanoTWICE** [16] is the first dataset proposed to apply the AD problem. It contains 45 nanofibrous material images with 1024×3696 pixels captured from a scanning electron microscope. The background of the image is a non-cyclical continuous texture, and the size of the defect varies.

**MVTec AD** [115] is currently the most common dataset for industrial AL, which contains 15 categories, each category has about 240 normal images for training and 100 defective images for testing. The original image resolution is between 700×700 and 1024×1024 pixels. Compared with the existing datasets that focus on texture defects, this dataset has ten objects and five texture types. The five categories cover different types of regular (carpet, grid) or random (leather, tile, wood) textures, while the remaining ten categories represent various types of objects. Some of these objects are rigid and have a fixed appearance (bottles, metal nuts), while others are deformable (cables) or include natural changes (hazelnuts). The test images of abnormal samples contain various defects, such as scratches, dents, structural differences, etc. There are a total of 73 different types of defects, with an average of about five for each category. More details of this dataset can be found in [115]. However, this dataset is well imaged and uniformly illuminated, while the image positions are fixed in some data types, making it a more idealized.

**BTAD** (beanTech Anomaly Detection) dataset has been recently released by Mishra et al. [85]. It contains 2830 images with three different classes. The resolutions of these three classes are 1600×1600, 600×600, and 800×600 pixels, respectively. Each class is composed of defect-free training and testing images, like the MVTec AD dataset, except that the defect types are not illustrated.

**Fabric dataset** [116] is from the automation laboratory sample database of Hong Kong University constructed by Tsang et al., which contains 256×256 fabric images belonging to three patterns: dot, star, and box-patterned fabrics. Each pattern has 25 defect-free and 25 defective samples. There are five types of defects that appear in the defective samples include the broken-end, hole, netting multiple, thick-bar, and thin-bar. All the defective fabric images have the corresponding ground-truth. This dataset is a classical textured dataset, which is often employed in fabric defect detection works.

**Textured dataset** is also created by MVTec company and first presented in the AE-SSIM [18]. This dataset contains two woven fabric textures. All images are of size 512×512 pixels that were acquired as single-channel gray-scale images.

Besides these, some datasets widely used for other supervised industrial vision tasks are also employed for AL, such as KolektorSDD [117], KolektorSDD2 [118], RSDD (Railway Surface Discrete Defects) [119], and magnetic tile (MT) defect

[120] datasets. Typically, the defect-free samples of the    training

**TABLE 9 Common datasets for industrial anomaly localization**

| Name | URL* | Description | Flaw |
|---|---|---|---|
| NanoTWICE [16] | http://www.mi.imati.cnr.it/ettore/NanoTWICE | 45 nanofibrous material images of size 1024 × 3696, non-cyclical continuous texture, only textural anomalies | Single continuous texture, too few samples |
| MVTec AD [115] | https://www.mvtec.com/company/research/datasets/mvtec-ad | 15 categories, with about 240 normal images for training and 100 defective images for testing in each category, Image size ranges from 700×700 to 1024×1024, textural anomalies and functional anomalies | Too consistent image illumination, serious alignment in some data sets, and too few functional anomalies |
| BTAD [85] | http://avires.dimi.uniud.it/papers/btad/btad.zip | 2830 images with 3 different classes, only textural anomalies | Serious alignment, too few defect types, too consistent image illumination |
| Fabric dataset [116] | https://ytngan.wordpress.com/codes | 3 categories, each having 25 defect-free and 25 defective samples, only textural anomalies | Uniform texture background, too consistent image illumination |
| Textured dataset [18] | https://www.mvtec.com/company/research/publications | 2 woven fabric textures with 512 × 512 pixels, only textural anomalies | Uniform texture, consistent image illumination |
| KolektorSDD [117] | https://www.vicos.si/resources/kolektorsdd | 347 images without defects and 52 images with defects, only textural anomalies | Only with one defect type-thin scratch |
| KolektorSDD2 [118] | https://www.vicos.si/resources/kolektorsdd2 | Over 3000 images containing several types of defects, rich defect scales, only textural anomalies | Uniform texture |
| RSDDs [119] | http://icn.bjtu.edu.cn/Visint/resources/RSDDs.aspx or, https://pan.baidu.com/s/1kM5Lh9-s2yImuQE2ks0dqg (password: lvth) | Images with sizes 160×160 and 55×55, only textural anomalies | Too few image samples and constant defect size |
| MT Defect [120] | https://github.com/abin24/Magnetic-tile-defect-datasets. | 5 types of metal defects, rich imaging scenes, only textural anomalies | A very few samples in each category, noise in labels |

set in these datasets are used for AL model training. Then the remaining samples, mainly defective, are utilized as test sets.

KolektorSDD [117] and KolektorSDD2 [118] are the datasets of metal surfaces collected in real industrial scenarios. KolektorSDD is relatively simple, with only one thin scratch defect. KolektorSDD2 is a real-world complex and well-annotated modern surface inspection dataset. It is constructed from color images of defective production projects, captured using a visual inspection system, and annotated by the company. The defects are annotated with fine-grained segmentation masks that vary in shape, size, and color, ranging from minor scratches to spots on large surface defects. The RSDD [119] and MT Defect [120] datasets are also frequently employed to evaluate anomaly localization. The RSDD dataset contains two types of datasets collected on real railroad tracks, and its texture and illumination vary significantly. The MT Defect dataset includes five types of defects: blowhole, break, uneven, fray, and crack, all of which have different resolutions. These defect images contain a series of industrial noises, such as the changes in light intensity, the defect's scale, and the texture's complexity. But these do not contain many types and variations of defects. In Table 9, we list multiple image datasets commonly used by the AL community, and specifically indicate their download links, descriptions, and flaws.

## B. Evaluation Criteria

AUROC (Area Under the Receiver Operating Characteristic curve), PRO (Per Region Overlap) score, and IoU (Intersection over Union) are the three primary evaluation metrics used in AL, as discussed below.

***Receiver Operating Characteristic curve (AUROC)***: The most frequent indication of AL is the area under the receiver operating characteristic curve (AUROC) of each pixel. The high AUROC value indicates that the model is less influenced by varied threshold settings while identifying anomalies. Normal pixels are identified as negative, whereas anomalous pixels are identified as positive. The true positive rate (TPR) is the percentage of pixels properly categorized as anomalous across the evaluated category, whereas the false positive rate (FPR) is the percentage of pixels incorrectly classified as abnormal, denoted as follows.

$$TPR = \frac{TP}{TP + FN} \; ; \; FPR = \frac{FP}{FP + TN} \; ; \quad (1)$$

where, TP, FP, TN, and FN denote true positive, false positive, true negative, and false negative, respectively.

The AUROC value may be determined by scanning across the range of thresholds and obtaining serial sorted values of TPR and FPR. However, in surface AD settings where only a tiny proportion of pixels are anomalous, the AUROC does not accurately reflect localization accuracy. The reason is that the false-positive rate is dominated by the very high number of non-anomalous pixels and is thus kept low despite false-positive detection. Therefore, despite achieving about 97% pixel AUROC, some state-of-the-art methods cannot produce fine-grained AL results. They often have more interference and introduce too much background area, as seen in the visualized anomaly maps. Despite this, AUROC is currently the dominant evaluation indicator used.

***Per Region Overlap (PRO) score***: Since the AUROC favors large anomalies, the PRO score is also employed for anomaly localization. For computing the PRO metric, anomaly scores are first thresholded to make a binary decision for each pixel whether an anomaly is present or not. For each connected component within the ground-truth, the relative overlap with the thresholded anomaly region is computed. Numerous approaches [21, 23, 50, 84, 88, 116] also use the PRO score to evaluate the performance of the model.

TABLE 10 Localization results (pixel AUROC %) of state-of-the-art AL methods on MVTec AD

| Category | Method | Mean | carpet | grid | leather | tile | wood | bottle | cable | capsule | hazelnut | metal nut | pill | screw | toothbrush | transistor | zipper |
|---|---|---|---|---|---|---|---|---|---|---|---|---|---|---|---|---|---|
| AE-based | AESc [36] | 86.0 | 91.0 | 95.0 | 87.0 | 79.0 | 84.0 | 88.0 | 84.0 | 93.0 | 93.0 | 62.0 | 85.0 | 95.0 | 93.0 | 78.0 | 90.0 |
|  | TrustMAE [41] | 93.9 | 98.5 | 97.5 | 98.1 | 82.5 | 92.6 | 93.4 | 92.9 | 87.4 | 98.5 | 91.8 | 89.9 | 97.6 | 98.1 | 92.7 | 97.8 |
|  | PixelSail [47] | 94.0 | 94.0 | 99.0 | 87.0 | 99.0 | 88.0 | 95.0 | 95.0 | 93.0 | 95.0 | 91.0 | 95.0 | 96.0 | 97.0 | 91.0 | 98.0 |
|  | UTAD [49] | 90.0 | - | - | - | - | - | - | - | - | - | - | - | - | - | - | - |
|  | MPAD [50] | 98.1 | 98.4 | 98.5 | 99.1 | 94.4 | 97.5 | 98.6 | 98.2 | 97.9 | 97.8 | 99.1 | 98.8 | 98.5 | 99.0 | 97.7 | 98.6 |
|  | MIGD [51] | 91.0 | - | - | - | - | - | - | - | - | - | - | - | - | - | - | - |
|  | AE-SSIM [18] | 86.2 | 87.0 | 94.0 | 78.0 | 59.0 | 73.0 | 93.0 | 82.0 | 94.0 | 97.0 | 89.0 | 91.0 | 96.0 | 92.0 | 80.0 | 88.0 |
|  | AE-$L_2$ [18] | 82.0 | 59.0 | 90.0 | 75.0 | 51.0 | 73.0 | 86.0 | 86.0 | 88.0 | 95.0 | 86.0 | 85.0 | 96.0 | 93.0 | 86.0 | 77.0 |
|  | RIAD [53] | 94.2 | 96.3 | 98.8 | 99.4 | 89.1 | 85.8 | 98.4 | 84.2 | 92.8 | 96.1 | 92.5 | 95.7 | 98.8 | 98.9 | 87.7 | 97.8 |
| Generative model-based | Vanilla VAE [55] | 82.3 | 62.0 | 85.6 | 83.5 | 52.0 | 69.9 | 89.4 | 81.6 | 90.7 | 95.1 | 86.1 | 87.9 | 92.8 | 95.3 | 85.1 | 77.5 |
|  | VE-VAE [57] | 86.1 | 78.0 | 73.0 | 95.0 | 80.0 | 77.0 | 87.0 | 90.0 | 74.0 | 98.0 | 94.0 | 83.0 | 97.0 | 94.0 | 93.0 | 78.0 |
|  | CAVAG [59] | 89.0 | - | - | - | - | - | - | - | - | - | - | - | - | - | - | - |
|  | VAE-grad [61] | 89.0 | 74.0 | 96.0 | 93.0 | 65.0 | 84.0 | 92.0 | 91.0 | 92.0 | 98.0 | 91.0 | 93.0 | 95.0 | 98.0 | 92.0 | 87.0 |
|  | AnoGAN [17] | 74.3 | 54.0 | 58.0 | 64.0 | 50.0 | 62.0 | 86.0 | 78.0 | 84.0 | 87.0 | 76.0 | 87.0 | 80.0 | 90.0 | 80.0 | 78.0 |
|  | CFLOW-AD [79] | 98.6 | 99.3 | 99.0 | **99.7** | 98.0 | 96.7 | 99.0 | 97.6 | 99.0 | 99.0 | 98.6 | 99.0 | 98.9 | 98.9 | 98.0 | 99.1 |
|  | Fastflow [28] | 98.5 | **99.4** | 98.3 | 99.5 | 96.3 | 97.0 | 97.7 | 98.4 | **99.1** | 99.1 | 98.5 | 99.2 | 99.4 | 98.9 | 97.3 | 98.7 |
| Deep feature embedding-based | S-T [21] | 93.9 | 93.5 | 89.9 | 97.8 | 92.5 | 92.1 | 97.8 | 91.9 | 96.8 | 98.2 | 97.2 | 96.5 | 97.4 | 97.9 | 73.7 | 95.6 |
|  | MKD [26] | 90.7 | 95.6 | 91.7 | 98.0 | 82.7 | 84.8 | 96.3 | 82.4 | 95.9 | 94.6 | 86.4 | 89.6 | 96.0 | 96.1 | 76.5 | 93.9 |
|  | STPM [81] | 97.0 | 98.8 | 99.0 | 99.3 | 97.4 | 97.2 | 98.8 | 95.5 | 98.3 | 98.5 | 97.6 | 97.8 | 98.3 | 98.9 | 82.5 | 98.5 |
|  | OE-SDN [82] | 93.0 | 96.0 | 97.0 | 85.0 | 85.0 | 82.0 | 95.0 | 84.0 | 97.0 | 98.0 | 93.0 | 93.0 | 97.0 | 98.0 | 89.0 | 91.0 |
|  | FAVAE [83] | 95.3 | 96.0 | 99.3 | 98.1 | 71.4 | 89.9 | 96.9 | 97.6 | 98.7 | 96.8 | 96.1 | 95.3 | 99.3 | 98.7 | **98.4** | 96.8 |
|  | SPADE [25] | 96.5 | 97.5 | 93.7 | 97.6 | 87.4 | 88.5 | 98.4 | 97.2 | 99.0 | 99.1 | 98.1 | 96.5 | 98.9 | 97.9 | 94.1 | 96.5 |
|  | PaDiM [23] | 97.5 | 99.1 | 97.3 | 99.2 | 94.1 | 94.9 | 98.3 | 96.7 | 98.5 | 98.2 | 97.2 | 95.7 | 98.5 | 98.8 | 97.5 | 98.5 |
|  | DFR [84] | 95.0 | 97.0 | 98.0 | 98.0 | 87.0 | 93.0 | 97.0 | 92.0 | 99.0 | 99.0 | 93.0 | 97.0 | 99.0 | 99.0 | 80.0 | 96.0 |
|  | SOE [87] | 98.2 | - | - | - | - | - | - | - | - | - | - | - | - | - | - | - |
|  | PatchCore [30] | 98.1 | 99.0 | 98.7 | 99.3 | 95.6 | 95.0 | 98.6 | 98.4 | 98.8 | 98.7 | 98.4 | 98.4 | 97.4 | 99.4 | 98.7 | 96.3 | 98.8 |
|  | SOM [88] | 97.8 | 98.9 | 98.4 | 99.1 | 94.8 | 94.4 | 98.3 | 98.2 | 98.7 | 98.4 | 98.0 | 98.0 | 99.1 | 98.5 | 95.3 | 98.7 |
|  | TF [89] | 96.5 | - | - | - | - | - | - | - | - | - | - | - | - | - | - | - |
|  | ALT [90] | 96.9 | 97.1 | 99.5 | 98.9 | 95.5 | 96.2 | 96.4 | 90.8 | 98.8 | 99.1 | 97.6 | 98.5 | 99.3 | 97.7 | 91.4 | 97.5 |
|  | FYD [92] | 98.2 | 98.5 | 96.8 | 99.2 | 96.8 | **99.6** | 98.3 | 97.5 | 98.6 | 98.7 | 98.2 | 97.3 | 98.7 | 98.9 | 98.1 | 98.2 |
| Self supervised learning-based | SMAI [95] | 89.0 | 88.0 | 97.0 | 86.0 | 62.0 | 80.0 | 86.0 | 92.0 | 93.0 | 97.0 | 92.0 | 92.0 | 96.0 | 96.0 | 85.0 | 90.0 |
|  | Cutpaste [97] | 96.0 | 98.3 | 97.5 | 99.5 | 90.5 | 95.5 | 97.6 | 90.0 | 97.4 | 97.3 | 93.1 | 95.7 | 96.7 | 98.1 | 93.0 | **99.3** |
|  | ANOSEG [98] | 97.0 | 99.0 | 99.0 | 98.0 | 98.0 | 98.0 | 99.0 | 99.0 | **99.0** | 90.0 | 99.0 | 99.0 | 94.0 | 91.0 | 96.0 | 96.0 | 98.0 |
|  | NSA [99] | 96.3 | 95.5 | 99.2 | 99.5 | **99.3** | 90.7 | 98.3 | 96.0 | 97.6 | 97.6 | 98.4 | 98.5 | 96.5 | 94.9 | 88.0 | 94.2 |
|  | DREAM [29] | 97.3 | 95.5 | **99.7** | 98.6 | 99.2 | 96.4 | **99.1** | 94.7 | 94.3 | 99.7 | **99.5** | 97.6 | 97.6 | 98.1 | 90.9 | 98.8 |
|  | InTra [37] | 96.6 | 99.2 | 98.8 | 99.5 | 94.4 | 88.7 | 97.1 | 91.0 | 97.7 | 98.3 | 93.3 | 98.3 | 99.5 | 98.9 | 96.1 | 99.2 |
|  | SSPCAB [101] | 97.2 | 95.0 | 99.5 | 99.5 | **99.3** | 96.8 | 98.8 | 96.0 | 93.1 | **99.8** | 98.9 | 97.5 | **99.8** | 98.1 | 87.0 | 99.0 |
|  | CPC-AD [104] | 82.0 | 74.0 | 80.0 | 94.0 | 82.0 | 82.0 | 89.0 | 84.0 | 72.0 | 81.0 | 76.0 | 77.0 | 65.0 | 81.0 | 90.0 | 95.0 |
|  | DLA [107] | 93.0 | 89.4 | 88.1 | 98.5 | 91.9 | 89.2 | 91.8 | 88.3 | 96.5 | 96.2 | 92.6 | 96.4 | 97.2 | 95.8 | 88.3 | 95.4 |
| One-class classification-based | Patch-SVDD [24] | 95.7 | 92.6 | 96.2 | 97.4 | 91.4 | 90.8 | 98.1 | 96.8 | 95.8 | 97.5 | 98.0 | 95.1 | 95.7 | 98.1 | 97.0 | 95.1 |
|  | SE-SVDD [113] | 97.5 | 98.9 | 97.2 | 98.7 | 92.3 | 95.1 | 98.6 | 97.7 | 98.5 | 98.0 | 98.3 | 96.7 | 98.6 | **99.3** | 97.2 | 97.9 |
|  | FCDD [114] | 92.0 | 96.0 | 91.0 | 98.0 | 91.0 | 88.0 | 97.0 | 90.0 | 93.0 | 95.0 | 94.0 | 81.0 | 86.0 | 94.0 | 88.0 | 92.0 |

In each column, the best result is marked **bold**

TABLE 11 Localization results (PRO score %) of state-of-the-art AL methods on MVTec AD

| Category | Method | Mean | carpet | grid | leather | tile | wood | bottle | cable | capsule | hazelnut | metal nut | pill | screw | toothbrush | transistor | zipper |
|---|---|---|---|---|---|---|---|---|---|---|---|---|---|---|---|---|---|
| AE-based | AE-SSIM [18] | 69.4 | 64.7 | 84.9 | 56.1 | 17.5 | 60.5 | 83.4 | 47.8 | 86.0 | 91.6 | 60.3 | 83.0 | 88.7 | 78.4 | 72.4 | 66.5 |
|  | PixelSail [47] | 50.0 | 47.0 | 89.0 | 80.0 | 36.0 | 53.0 | 52.0 | 40.0 | 31.0 | 54.0 | 36.0 | 24.0 | 47.0 | 69.0 | 8.0 | 82.0 |
|  | MPAD [50] | **95.5** | 92.7 | 97.9 | **99.2** | 88.8 | **96.2** | 95.3 | **96.7** | **97.8** | **97.8** | 88.8 | 96.1 | 98.3 | 94.4 | 95.0 | 97.0 |
| Generative model-based | Vanilla VAE [55] | 64.2 | 61.9 | 40.8 | 64.9 | 24.2 | 57.8 | 70.5 | 77.9 | 77.9 | 77.0 | 57.6 | 79.3 | 66.4 | 85.4 | 61.0 | 60.8 |
|  | CFLOW-AD [79] | 94.6 | **97.7** | 96.1 | 99.4 | 94.3 | 95.8 | **96.8** | 93.5 | 93.4 | 96.7 | 91.7 | 95.4 | 95.3 | 95.1 | 91.4 | 96.6 |
| Deep feature embedding based | S-T [21] | 91.4 | 87.9 | 95.2 | 94.5 | 94.6 | 91.1 | 93.1 | 81.8 | 96.8 | 96.5 | 94.2 | 96.1 | 94.2 | 93.3 | 66.6 | 95.1 |
|  | STPM [81] | 92.1 | 95.8 | 96.6 | 98.0 | 92.1 | 93.6 | 95.1 | 87.7 | 92.2 | 94.3 | 94.5 | **96.5** | 93.0 | 92.2 | 69.5 | 95.2 |
|  | PaDiM [25] | 91.7 | 94.7 | 86.7 | 97.2 | 75.9 | 87.4 | 95.5 | 90.9 | 93.7 | 95.4 | 94.4 | 94.6 | 96.0 | 93.5 | **97.4** | 92.6 |
|  | PaDiM [23] | 92.1 | 96.2 | 94.6 | 97.8 | 86.0 | 91.1 | 94.8 | 88.8 | 93.5 | 92.6 | 85.6 | 92.7 | 94.4 | 93.1 | 84.5 | 95.9 |
|  | DFR [84] | 91.0 | 93.0 | 93.0 | 97.0 | 79.0 | 91.0 | 93.0 | 81.0 | 97.0 | 97.0 | 90.0 | 96.0 | 96.0 | 93.0 | 79.0 | 90.0 |
|  | SOE [87] | 94.2 | - | - | - | - | - | - | - | - | - | - | - | - | - | - | - |
|  | PatchCore [30] | 93.5 | 96.6 | 95.9 | 98.9 | 87.4 | 89.6 | 96.1 | 92.6 | 95.5 | 93.9 | 91.3 | 94.1 | 97.9 | 91.4 | 83.5 | **97.1** |
|  | SOM [88] | 93.3 | 95.5 | 95.3 | 97.7 | 81.3 | 88.2 | 94.7 | 93.4 | 93.4 | 95.1 | 93.6 | 96.5 | 96.0 | 90.7 | 91.6 | 95.9 |
|  | TF [89] | 88.7 | - | - | - | - | - | - | - | - | - | - | - | - | - | - | - |
|  | ALT [90] | 94.0 | 91.3 | **99.0** | 96.8 | 90.1 | 92.5 | 94.3 | 88.9 | 97.1 | **98.0** | 90.3 | 94.0 | **98.9** | 96.3 | 88.4 | 94.0 |
|  | FYD [92] | 91.8 | - | - | - | - | - | - | - | - | - | - | - | - | - | - | - |
| Self supervised learning based | NSA [99] | 91.0 | 85.0 | 96.8 | 98.7 | **95.3** | 85.3 | 92.9 | 89.9 | 91.4 | 93.6 | **94.6** | 96.0 | 90.1 | 90.7 | 75.3 | 89.2 |
| One-class classification based | SE-SVDD [113] | 92.3 | 96.1 | 94.3 | 96.2 | 87.5 | 90.7 | 93.9 | 87.9 | 93.3 | 93.7 | 93.5 | 93.2 | 93.3 | 95.3 | 93.1 | 85.5 |

In each column, the best result is marked **bold**

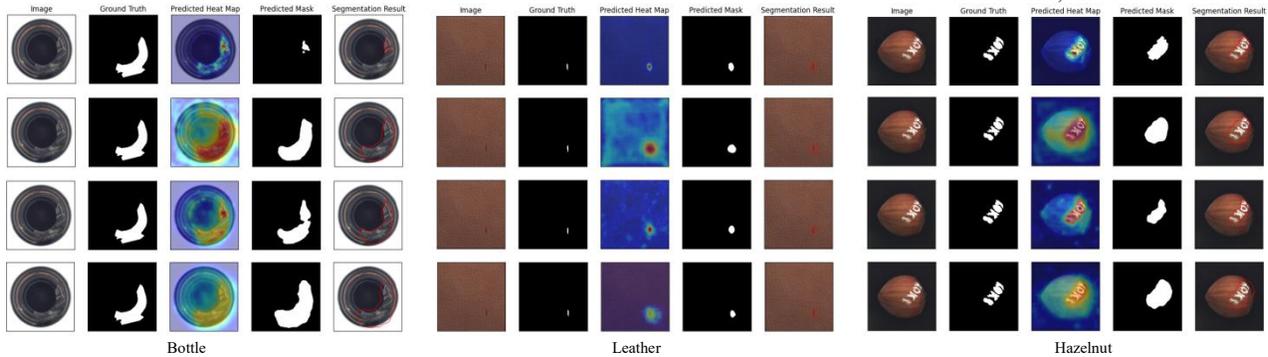

Fig. 13. Qualitative results on MVTec AD samples of STPM [81], PatchCore [30], PaDiM [23], and CFLOW-AD [79], row-wise.

***Intersection-over-Union (IoU)***: AL can be treated as a segmentation task similar to that in supervised learning. The IoU, as the primary metric for segmentation tasks, can equally justly be used to evaluate the performance of AL. Currently, only very few works use this evaluation approach, such as [49], [59], and [98].

## C. Performance Comparison

***Performance on MVTec AD datasets***: Tables 10 and 11 summarize the performance of the contemporary AL methods (published mainly from the year 2017 to 2021) on the MVTec AD dataset. We observed that most of these approaches achieved baseline performance with assistance from AE. A few attempts are dedicated to designing more powerful modules, such as image inpainting and GAN. The pixel AUROC on the MVTec AD dataset has been up to 94.2% achieved by RIAD [53]. Nevertheless, experimental results demonstrated these pure AE-based reconstructions or generative methods are hardly capable of performing sufficiently well on the MVTec AD dataset.

In contrast, deep feature embedding-based methods have quickly demonstrated their strengths in AL. The reported results in the past papers show that three typical feature compared methods, S-T [21], SPADE [25], and DFR [84], achieved overall 93.9%, 96.5%, and 95.0% pixel AUROC on the MVTec AD dataset, respectively. Starting from the generic feature modeling method [23], feature embedding-based methods improve steadily when introduced with more effective strategies, e.g., feature selection into semi-orthogonal embedding [87], attention strategy [23, 43], KNN with the memory bank [30], self-organizing feature [88] and aligning feature [92]. As a result, most approaches yielded about a pixel AUROC of 93% and a PRO score of 91% on the MVTec AD dataset. Moreover, CFLOW-AD [79], combined with a novel generative network, outperformed other state-of-the-art models and achieved the best pixel AUROC on MVTec AD so far. On the other hand, MPAD [50], combined with pre-trained features, surpassed other state-of-the-art models and achieved the best PRO score on MVTec AD so far. Here, in Fig. 13, we present the visualization of AL results of four typical feature embedding methods over MVTec AD, including STPM [81], PatchCore [30], PaDiM [23], and CFLOW-AD [79]. These results were obtained using a standard image AL library Anomalib [125], maintained by Intel corporation.

Self-supervised learning-based methods can learn visual features from unlabeled images and be embedded into the above network structure as an additional module. Such methods, e.g., ANOSEG [98], NSA [99], and DRAEM [29], can achieve better results compared with original AE-based methods. Moreover, contrast learning-based methods [92, 107] demonstrate very competitive performance due to the discriminative information of the anomaly regions, compared to image reconstruction or pre-trained features. One-class classification-based methods are usually time-consuming and obtain inaccurate localization results, especially the computational time of cropping local patches and extracting individual local features. However, some methods include more complicated feature comparison procedures, e.g.,

Patch-SVDD [24] and SE-SVDD [113] are designed in a unified pipeline to improve the localization performance.

In summary, the deep learning-based AL methods can obtain a relatively satisfactory result on the MVTec AD dataset by adopting different strategies. In particular, three datasets among the 15 datasets have not been overcome by the majority of methods; those are the Tile, Wood, and Transistor datasets. Tile and wood are typical texture datasets that contain multi-scale and multi-type defects, and major methods currently do not achieve 95% AUROC. The Transistor dataset has a missing defect type containing high-level semantic information. In this dataset, it treats all ranges of the missing as ground-truth. Hence major current methods also do not achieve ideal performances.

***CNN vs. ViT vs. NF***: We also analyze the performance of the MVTec AD dataset using different network modules under a deep feature embedding-based approach. Results in pixel AUROC (%) of representative methods are shown in Table 12. The two best algorithms, CFLOW-AD [79] and Fastflow [28], both employed the NF module training method. UTRAD [124] and InTra [27] are two approaches that leveraged ViT by utilizing transformer layers to build the reconstruction model. Other algorithms used a simple CNN module. ViT can capture a wide variety of visual areas through an attention mechanism, whereas NF directly estimates the probability of a normal sample. Table 12 comprehends that combining ViT or NF can significantly improve the localization accuracy. As a matter of fact, more attention is encouraged to use NF and ViT in the future.

TABLE 12 Comparison of different network modules on MVTec-AD

| Methods | | Venue | Key points | Performance |
|---|---|---|---|---|
| Vanilla CNN | S-T [21] | IJCV-2021 | Knowledge distillation | 93.9 |
| | MKD [26] | CVPR-2021 | Knowledge distillation | 90.7 |
| | SPADE [25] | ArXiv-2020 | Feature modeling | 96.5 |
| | DFR [84] | NeuroComp2021 | Feature modeling | 95.0 |
| | FAVAE [83] | ArXiv-2020 | Feature modeling | 95.3 |
| ViT | InTra [27] | ArXiv-2021 | Image inpainting | 96.6 |
| | UTRAD [124] | Neural Net-2021 | Feature modeling | **96.7** |
| NF-based | CFLOW-AD [79] | WACV-2022 | Feature modeling | **98.6** |
| | Fastflow [28] | ArXiv-2021 | Feature modeling | 98.5 |

In the last column, the best result is marked **bold**

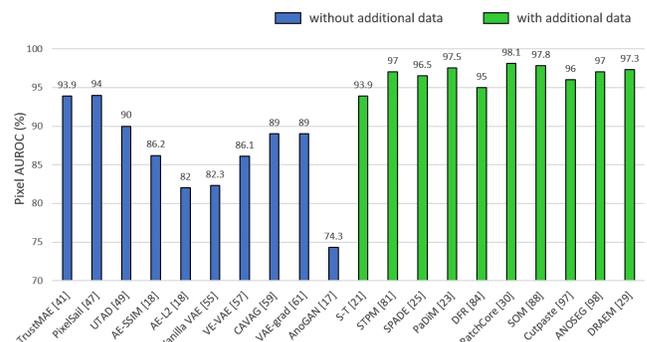

Fig. 14. Comparison among state-of-the-art AL methods with/without additional data on the MVTec AD dataset.

***Additional data vs. without additional data***: AL in industrial scenarios is a very challenging problem. It is difficult to achieve good results by relying solely on image-level data reconstruction or generation. Most existing approaches take assistance from additional data used in large pre-trained models, data synthesis by self-supervised means, or designing proxy tasks. In Fig. 14, we present a bar graph for easier performance comparison of models with and without additional data usage over MVTec AD. It can be observed that major methods employing additional data outperformed the methods that did not use extra data.

***Run-time analysis***: The real-time characteristic of the AL in industrial images is a special feature. Hence, in Table 13, we provide the run-time analysis of some major AL methods. As shown in this table, the AL methods based on deep feature embedding touch 20 fps. In particular, CFLOW-AD-WRN50 [79] achieved 27 fps, demonstrating its ability for defect detection in real-time.

TABLE 13 Inference speed (fps) of various AL methods on MVTec-AD

| Methods | Inference speed (fps) |
|---|---|
| AnoGAN [17] | 0.02 |
| GANomaly [67] | 9.1 |
| Skip-GANomaly [35] | 7.9 |
| DifferNet [78] | 2.04 |
| PaDiM-Resnet18 [23] | 4.4 |
| SPADE-WRN50 [25] | 0.1 |
| CFLOW-AD-WRN50 [79] | **27** |
| DFR [84] | 20 |
| Patch Core-WRN50 [30] | 5.88 |
| FastFlow-WRN50 [28] | 21.8 |

In the last column, the best result is marked **bold**

## V. Conclusion and outlook

This paper has highlighted recent achievements in industrial AL using deep learning. Here, we also provided some structural taxonomy for various methods based on their roles for AL, analyzed their advantages and limitations, summarized existing popular industrial AL datasets, and discussed performance for the most representative approaches. Despite significant progress, several issues remain unresolved. This section will highlight these issues and present some possible future research directions. We anticipate that this study not only improves an academic understanding of industrial AL, but also encourages future research efforts.

***Functional anomalies:*** From the strengths and weaknesses mentioned in the above tables, it can be observed that the localization effects of many methods drop significantly on some specific data sets. For example, the disadvantage of DFR [84] is the poor performance on transistor data sets (refer to Tables 6, 10). This is because most of the data sets shown in Table 10 are textural defects, such as scratches and dents, rather than functional anomalies. Functional anomalies violate underlying constraints, e.g., a permissible object in an invalid location or the absence of a required object. In industrial scenes, both types are equally important. At present, there is already a method by Bergmann et al. [126] to jointly detect the textural and functional anomalies. However, the research on functional defects will be an important direction in the future.

***Releasing rich AL datasets***: Compared to real industry scenarios, public anomaly location datasets are not yet large or rich enough. More complex datasets with changing imaging conditions such as lighting, perspective, proportion, shadow, blur, etc., should be available to evaluate the effect of the AL algorithm more objectively. The existing MVTec AD has single imaging, relatively good image quality, and alignment in some categories. Some existing AL methods even exploit this property for performance enhancement. Despite the promising results achieved, these methods cannot be adapted to real complex industrial scenarios. Therefore, it is necessary to have some realistic and rich industrial AL datasets.

***Vision transformer-based method***: The ViT-based methods currently dominate the field of computer vision since their superior performance. Some ViT-based works [27, 124, 79] have also been proposed to solve the AL problem. ViT has particular advantages in long-distance feature modeling. Comprehensively considering multi-scale anomalous regions is a direction that ViT can improve. Moreover, the best framework for AL is the generation model based on NF. Therefore, the combination of ViT and NF also has been an important direction.

***Meaningful model evaluation***: As stated above, there is an overlap between the high pixel-AUROC value and the fine-grained localization performance, which may cause the model validity problem. Many methods still utilize the pixel-AUROC evaluation metric, but the visualization results of the AL do not perform well. Future works are recommended to consider the problem of fine boundary when building their models or choose the IoU metric for model evaluation.

***Accurate anomaly types***: The types of anomalies in real industrial scenarios are diverse, and the importance of different anomaly types varies. This problem challenges the classical paradigm of AD or localization and requires the development of learning methods that can discriminate between anomaly types. There are already methods [122] to cluster anomaly types and group anomaly data into semantically consistent categories, but this is only a start.

***Unsupervised 3D anomaly localization***: With the spread of 3D sensors, an increasing number of defect detection tasks in industrial scenarios are moving from 2D to 3D scenarios. Correspondingly, AL in 3D scenes will be a trend for the sake of development. Recently, MVTec company made a 3D AD/ AL dataset publicly available in late 2021 [123]. Therefore, we believe 3D AD/AL constitutes a relevant future direction.